\documentclass[letterpaper, 10 pt, conference]{ieeeconf}

\IEEEoverridecommandlockouts                              

\usepackage{amsmath,amsfonts}
\usepackage{algorithmic}
\usepackage{tikz}
\usepackage{algorithm}
\usepackage{array}
\usepackage{xcolor}
\usepackage[caption=false,font=normalsize,labelfont=sf,textfont=sf]{subfig}
\usepackage{textcomp}
\usepackage{csquotes}
\usepackage{url}
\usepackage{graphicx}
\usepackage{cite}
\usepackage[T1]{fontenc}
\usepackage{hyperref}
\newtheorem{theorem}{Theorem}
\newtheorem{prop}{Proposition}
\newtheorem{remark}{Remark}

\title{ Closing Speed Computation using Stereo Camera and Applications in Unsignalized T-Intersection}

\author{Gautam Kumar$^{1}$ and Ashwini Ratnoo$^{2}$
\thanks{$^{1}$ Gautam Kumar is a Ph.D. student in the Department of Aerospace Engineering, Indian Institute of Science, Bangalore 560012, India.
        {\tt\small gautamkumar1@iisc.ac.in}}%
\thanks{$^{2}$ Ashwini Ratnoo is an Associate Professor (AcP) in the Department of Aerospace Engineering, Indian Institute of Science, Bangalore 560012, India.
        {\tt\small ratnoo@iisc.ac.in}}%
}

\begin{document}
\maketitle
\begin{tikzpicture}[remember picture, overlay]
    \node[anchor=north, yshift=-10pt] at (current page.north) {
        \parbox{\textwidth}{\centering\footnotesize
        This work has been submitted to the IEEE for possible publication. Copyright may be transferred without notice, after which this version may no longer be accessible.}
    };
\end{tikzpicture}

\begin{abstract}

This letter presents a conflict resolution strategy for an autonomous vehicle mounted with a stereo camera approaching an unsignalized T-intersection. A mathematical model for uncertainty in stereo camera depth measurements is considered and an analysis establishes the proposed adaptive depth sampling logic which guarantees an upper bound on the computed closing speed. Further, a collision avoidance logic is proposed that utilizes the closing speed bound and generates a safe trajectory plan based on the convex hull property of a quadratic B\'ezier curve-based reference path. Realistic validation studies are presented with neighboring vehicle trajectories generated using Next Generation Simulation (NGSIM) dataset.

\end{abstract}

\section{Introduction}

Recent advances in autonomous vehicle technology have opened up many challenges in trajectory planning \cite{zeng2019novel} and intersection management \cite{chouhan2018autonomous}. Autonomous operation of an ego vehicle has the potential to increase passenger safety and improve traffic efficiency \cite{garg2023can}. This comes with the challenge of generating a safe trajectory planning solution which considers the environment surrounding the ego vehicle typically perceived by onboard sensors such as vision systems, radar, and lidar.

In urban traffic scenarios, the problem of trajectory planning for an ego vehicle is addressed in different driving conditions like roundabouts \cite{wu2020efficient}, intersections \cite{qin2023traffic} and parking maneuvers \cite{upadhyay2018point}. An unsignalized T-intersection is characterized by a junction of two roads with a \enquote{STOP} sign or absence of any sign, and is one of the frequently encountered traffic situations, such as in lane-exit, driveways, and private entrances. Conflict-free trajectory planning at unsignalized T-intersection requires formulating a decision-making strategy based on the position and velocity information of the neighboring vehicles \cite{hang2022driving}.

\subsection{Related Works}

In \cite{aksjonov2021rule}, the ego vehicle utilizes intersection geometry and neighboring vehicles' path to identify potential collision locations. The collision detection therein is based on the closing speed and depth information of the observed neighboring vehicle, and upon such detection the ego vehicle stops until the neighboring vehicle has not crossed the intersection. A risk assessment-based driving strategy avoiding conflict at a T-intersection is proposed in \cite{park2023occlusion} where the ego vehicle resolves conflict through velocity profile planning. In \cite{chen2019autonomous}, a motion planning scheme considers the position and velocity information of the neighboring vehicles and generates a collision-free circular arc trajectory for the ego vehicle at a T-intersection. The aforementioned approaches assume accurate neighboring vehicle position and velocity information while formulating conflict-free trajectories. In \cite{fu2023efficient}, the conflict resolution strategy considers the propagation of uncertainty in the predicted position of the surrounding traffic participants using a Gaussian distribution and a Kalman filter. While that approach takes into account factors such as driver intentions and driving environments, the uncertainty in position and velocity measurements based on sensor characteristics is neglected.

Owing to the rapid developments in computer vision algorithms \cite{yang2021driving} and cost-effectiveness \cite{feng2020deep}, vision-based sensors are commonly used perception systems in autonomous vehicles. Because of the associated uncertainty in position measurement using vision sensors \cite{skibicki2020influence}, accurate trajectory prediction of the neighboring vehicles is necessary for generating safe and reliable trajectory solutions for the ego vehicle. As against monocular cameras, stereo cameras are widely used in automotive applications as they offer reliable depth information regardless of the road scenarios \cite{kemsaram2020stereo}. Common closing speed computation approaches using a stereo vision system involve the average distance travelled between two successive frames \cite{jalalat2016vehicle,yang2019vehicle}, or the average distance travelled in a fixed time interval \cite{gorodnichev2020automated}. Further, closing speeds computed using consecutive measurements taken over short sections lead to large speed errors \cite{fernandez2021vision}. This is primarily because the depth measurement accuracy of a stereo camera is significantly influenced by the resolution errors and camera parameters \cite{li2021geometrical}. The approaches in \cite{jalalat2016vehicle,yang2019vehicle,gorodnichev2020automated} do not consider the underlying uncertainty in stereo camera depth measurements, leading to significant deterioration of accuracy in computing closing speed which is one of the vital elements in decision-making at the intersection. Our work is motivated by developing a depth sampling logic that is based specifically on the uncertainty characteristics of stereo vision, leading to deterministic bounds on closing speed uncertainty, and hence paving the way for efficient conflict resolution.

\subsection{Our Contribution}
In this work, we consider an autonomous vehicle mounted with an on-board stereo camera system that provides the depth information of the neighboring vehicle. The main contributions are as follows:
\begin{enumerate}
    \item Considering an uncertainty model in depth information of the neighboring vehicle, an adaptive depth sampling logic is deduced to impose an upper bound on the computed closing speed.
    \item A quadratic B\'ezier curve-based path is considered for lane-exit maneuver at the T-intersection. Using the convex hull formed from the control points of the B\'ezier curve and the upper bound on the closing speed, a deterministic conflict resolution algorithm is proposed.
    \item The proposed solution is validated in a dynamic traffic scenario where the neighboring vehicles' trajectory is governed by realistic Next Generation Simulation (NGSIM) data.
\end{enumerate}
The rest of the paper is organized as follows. The problem scenario is discussed in Section \ref{sec:prob}. Section \ref{sec:vision} discusses the proposed depth sampling logic. Section \ref{sec:conflict} discusses the lane-exit path planning and the proposed conflict resolution strategy. Section \ref{sec:simulation} presents validation studies followed by concluding remarks in Section \ref{sec:conclusion}.

\section{Problem Statement}\label{sec:prob}

\begin{figure}[!hbt]
    \centering
    \includegraphics[trim={.1cm 0cm 1.1cm 3.6cm},clip,width=\columnwidth,keepaspectratio]{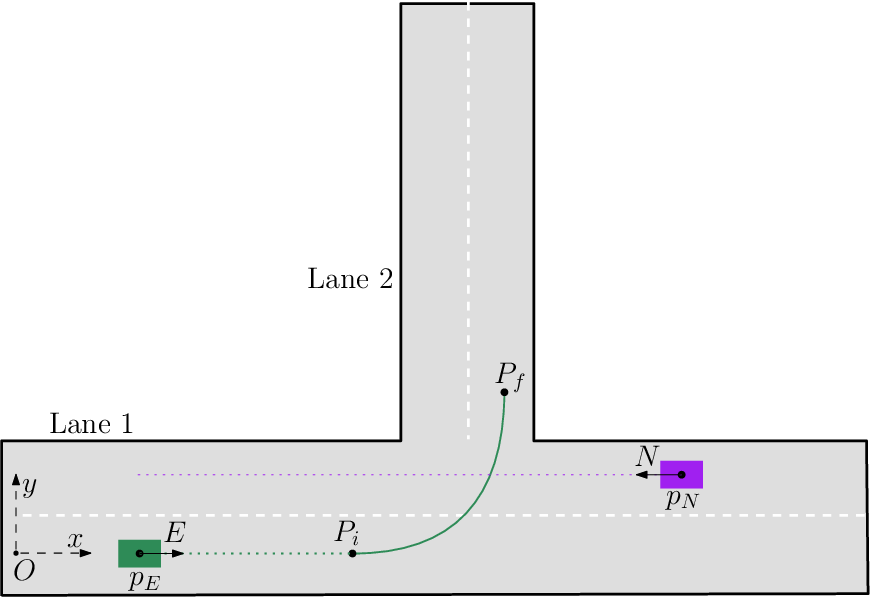}
    \caption{Lane exit scenario}
    \label{fig:prob_scenario}
\end{figure}

Consider an autonomous ego vehicle $E$ approaching a T-intersection as shown in Fig. \ref{fig:prob_scenario}. The ego vehicle needs to devise a decision-cum-planning strategy for performing a lane exit maneuver from Lane 1 (starting at point $P_i$) to Lane 2 (arriving at point $P_f$) while avoiding collision with a neighboring vehicle $N$ approaching the T-intersection in a direction opposite to the ego vehicle's motion. The neighboring vehicle is assumed to be moving continuously within its prescribed lane. Using the point mass representation, the reference position of the ego vehicle is indicated by $p_E =[x_E,y_E]^T$ and that for the neighboring vehicle is denoted by $p_N = [x_N,y_N]^T$ as expressed in a fixed frame.
The ego vehicle motion is governed by 
\begin{align}
\Dot{p}_E=
    \begin{bmatrix}
           \Dot{x}_E(t) \\
           \Dot{y}_E(t) \\
         \end{bmatrix} = 
         \begin{bmatrix}
           V_E \cos{\alpha_E} \\
           V_E \sin{\alpha_E} \\
         \end{bmatrix}
\end{align}
where $\alpha_E$ and $V_E$ are the heading angle and forward velocity of the ego vehicle, respectively. The instantaneous position of the neighboring vehicle as measured by a stereo camera onboard the ego vehicle is given by
\begin{equation}
\Tilde{p}_N = \begin{bmatrix}
           \Tilde{x}_N(t) \\
           \Tilde{y}_N(t) \\
         \end{bmatrix}=p_E + \mathcal{R}_{\alpha} \begin{bmatrix}
           x_m(t) \\
           y_m(t) \\
         \end{bmatrix}
\end{equation}
where $x_m$ and $y_m$ are the depth and relative lateral position of the neighboring vehicle measured by the stereo camera, and $\mathcal{R}_{\alpha}(\alpha_E) = \begin{bmatrix}
          \cos{\alpha_E}& -\sin{\alpha_E} \\
           \sin{\alpha_E} &  \cos{\alpha_E} 
         \end{bmatrix}$ transforms the position of the neighboring vehicle from the ego vehicle's body frame to the inertial frame. With $[x_m,y_m]^T$ as the only sensor measurement, the problem is to determine appropriate inputs $(V_E(t),\alpha_E(t))$ to the ego vehicle for executing a lane exit maneuver connecting $P_i$ and $P_f$ while avoiding collision with the neighboring vehicle by satisfying
\begin{align}\label{eq:ds}
\begin{aligned}
  (V_E(t),\alpha_e(t)) &: 
    \lVert p_E(t) -p_N(t) \rVert \geq d_s \forall\ t_i\leq t \leq t_f 
\end{aligned}
\end{align}
where $d_s$ is the minimum safety separation between two vehicles, and $t_i$ and $t_f$ correspond to the instants when the ego vehicle is at points $P_i$, and $P_f$, respectively.

\section{closing speed Computation with Uncertainty in Position}\label{sec:vision}
\subsection{Stereo Camera Position Uncertainty Model}
A stereo vision system uses a pair of cameras to perceive the position information of image features. While the lateral position can be deduced using the lane geometry and its associated error is not significant \cite{sankowski2017estimation}, the error in depth measurement using a stereo camera increases with depth as shown by experimental studies carried out in \cite{ortiz2018depth,cabrera2018versatile}. In \cite{cabrera2018versatile}, a quadratic polynomial is shown to approximate that trend with reasonable accuracy. Accordingly, the depth measurement error function is considered as
\begin{equation}\label{eq:error_model}
    f(x) = x_m-x = \beta_1x^2+\beta_2x+\beta_3
\end{equation}
where $x_m$ is the measured depth, $x$ is the computed depth, and ($\beta_1,\beta_2,\beta_3$) are curve fitting parameters. 
\begin{remark}\label{rem:fx}
    Owing to the systematic and random errors in camera calibration as noted in \cite{sankowski2017estimation}, the measured depth is higher than the computed depth, that is, $x_m>x$. This implies that $\beta_3>0$. Further, the difference $(x_m-x)$ becomes large at higher values of $x$, which implies that $f(x)$ is monotonically increasing at $x\gg 0$. This results in $\beta_1>0$.
\end{remark}
\begin{prop}\label{prop:xm}
For $x_m>\beta_3$, the roots of \eqref{eq:error_model} are real and the positive root is given by
\begin{equation}\label{eq:reldepth}
    x=\frac{-(\beta_2+1) + \sqrt{(\beta_2+1)^2-4 \beta_1\left(\beta_3-x_m\right)}}{2 \beta_1}
\end{equation}
\end{prop}
\begin{proof}
Solving for the depth $x$ using \eqref{eq:error_model},
\begin{equation}\label{eq:roots}
    x=\frac{-(\beta_2+1) \pm \sqrt{(\beta_2+1)^2-4 \beta_1\left(\beta_3-x_m\right)}}{2 \beta_1}
\end{equation}
For $x$ to be real in \eqref{eq:error_model}, 
\begin{equation}\label{eq:realroot}
    x_m \geq \beta_3 -\dfrac{(\beta_2+1)^2}{4\beta_1}
\end{equation}
Further, to ensure one of the roots in \eqref{eq:roots} is positive,
\begin{align}
    &\frac{-(\beta_2+1)+ \sqrt{(\beta_2+1)^2-4 \beta_1\left(\beta_3-x_m\right)}}{2 \beta_1}\geq 0\\
   \implies& -4 \beta_1\left(\beta_3-x_m\right)\geq 0~(\text{since } \beta_1>0)\\
   \implies&x_m \geq \beta_3 \label{eq:posroot}
\end{align}
\vspace{0.1cm}
Using $\beta_1>0$, for both \eqref{eq:realroot} and \eqref{eq:posroot} to satisfy, 
\begin{align}\label{eq:xm3}
    x_m\geq \beta_3.
\end{align}
\end{proof}

Further, to analyze the uncertainty in the computed depth, it is crucial to quantify how accurate is the curve fit model introduced in \eqref{eq:error_model}. The \textit{coefficient of determination} or $R^2$ is a measure of how well a curve model fits the data, and it is the quotient of explained variation (sum of squares due to regression) and the total variation \cite{cod}, that is,
\begin{equation}
    R^2  =\dfrac{\sum\limits_{i=1}^n (\hat{z}_i-\bar{z})^2}{\sum\limits_{i=1}^n(z_i-\bar{z})^2}
\end{equation}
where $z_i$ is the $i$th measured data, $\hat{z}_i$ is obtained from the curve-fitting polynomial, and $\bar{z}$ is the mean of all data points. 

\begin{remark}\label{rem:cod}
    The coefficient of determination satisfies $R^2 \in (0,1)$. If $R^2$ is close to 1, the curve-fitting polynomial traces the data points accurately.
\end{remark}

\begin{prop}\label{proposition:bounds}
If $R^2$ is the coefficient of determination of the error model in \eqref{eq:error_model}, the upper and lower bounds on computed depth $x$, denoted by $x_u$ and $x_l$, respectively, are given by 
\begin{align}
   &\begin{aligned}\label{eq:up_depth}
        &x_u=C_{0u} + \sqrt{C_{1u}+C_{2u}x + C_{3u}x^2}, & \forall x \in [0,\infty)
   \end{aligned}  ,\\
   &\begin{aligned}\label{eq:low_depth}
       x_l&=C_{0l} + \sqrt{C_{1l}+C_{2l}x + C_{3l}x^2}, ~ \\ &\forall x \in \left[\frac{-(\beta_2+1) + \sqrt{(\beta_2+1)^2-4 \beta_1\beta_3 U_f}}{2 \beta_1},\infty \right),
   \end{aligned}
\end{align}
where $C_{ju}$ and $ C_{jl},~ \forall j =\{0,1,2,3\}$ are constants and are described as 
\begin{align}
   & \begin{aligned}\label{eq:coeffup}
   &C_{0u} = -\dfrac{\beta_2+1-\beta_2U_f}{2\beta_1(1-U_f)},C_{1u} = C_{0u}^2+\dfrac{\beta_3U_f}{\beta_1(1-U_f)},\\
   &C_{2u} = \dfrac{\beta_2+1}{\beta_1(1-U_f)}, C_{3u} = \dfrac{1}{1-U_f}
    \end{aligned}\\
   & \begin{aligned}\label{eq:coefflow}
   &C_{0l} = -\dfrac{\beta_2+1+\beta_2U_f}{2\beta_1(1+U_f)},C_{1l} = C_{0l}^2-\dfrac{\beta_3U_f}{\beta_1(1+U_f)},\\
   &C_{2l} = \dfrac{\beta_2+1}{\beta_1(1+U_f)}, C_{3l} = \dfrac{1}{1+U_f}
    \end{aligned}
\end{align}
\end{prop}
\begin{proof}
Given the coefficient of determination factor $R^2$, the uncertainty factor $U_f = 1-R^2$ \cite{bai2022highly}. Using \eqref{eq:error_model}, the upper and lower bounds on $x$ satisfy 
\begin{align}
&x_m-x_u=(1-U_f) \left(\beta_1 x_u^2+\beta_2 x_u+\beta_3\right), \label{eq:upper_bound}\\
&x_m-x_l=(1+U_f) \left(\beta_1 x_l^2+\beta_2 x_l+\beta_3\right). \label{eq:lower_bound}
\end{align}
Substituting for $x_m$ from \eqref{eq:error_model} into \eqref{eq:upper_bound} and \eqref{eq:lower_bound},
\begin{align}
&\begin{aligned}
     (1-U_f)\beta_1 x_u^2&+(\beta_2+1-\beta_2U_f)x_u+\beta_3(1-U_f)-\\&\beta_1x^2-(\beta_2+1)x-\beta_3=0,
\end{aligned}\label{eq:quad_up}\\
&\begin{aligned}
     (1+U_f)\beta_1x_l^2&+(\beta_2+1+\beta_2U_f)x_l+\beta_3(1+U_f)-\\&\beta_1x^2-(\beta_2+1)x-\beta_3=0.
\end{aligned}\label{eq:quad_low}
\end{align}

Solving for $x_u$ and $x_l$ in \eqref{eq:quad_up} and \eqref{eq:quad_low}, respectively,
\begin{align}
      x_u&=C_{0u} \pm \sqrt{C_{1u}+C_{2u}x + C_{3u}x^2},\label{eq:quad_up1}\\
    x_l&=C_{0l} \pm \sqrt{C_{1l}+C_{2l}x + C_{3l}x^2}\label{eq:quad_low1}.
\end{align}
Further, $R^2 \in (0,1) \implies U_f \in (0,1)$. Using \eqref{eq:error_model} and \eqref{eq:coeffup},
\begin{align}\label{eq:quad_up2}
    C_{1u}+C_{2u}x + C_{3u}x^2 = C_{0u}^2 + \dfrac{x_m-\beta_3(1-U_f)}{\beta_1(1-U_f)}
\end{align}
Using $\beta_1,\beta_3>0$ in \eqref{eq:quad_up2},
\begin{align}
    &C_{1u}+C_{2u}x + C_{3u}x^2\geq C_{0u}^2,& \forall x_m \in [\beta_3,\infty). \label{eq:quad_up3}
\end{align}

Using \eqref{eq:quad_up3}, one of the roots in \eqref{eq:quad_up1} is positive and is given by
\begin{align}
    &x_u=C_{0u} + \sqrt{C_{1u}+C_{2u}x + C_{3u}x^2},~  \forall x_m \in [\beta_3,\infty)
\end{align}
where $x = 0$ when $x_m = \beta_3$ and as $x_m \rightarrow \infty ,x \rightarrow \infty $ using \eqref{eq:reldepth}, that is,
\begin{equation}\label{eq:quad_up4}
    x_m \in [\beta_3,\infty) \implies x \in [0,\infty].
\end{equation}
Similarly, we analyze the roots of \eqref{eq:quad_low1}. Using \eqref{eq:error_model} and \eqref{eq:coefflow},
\begin{align}\label{eq:quad_low2}
    C_{1l}+C_{2l}x + C_{3l}x^2 = C_{0l}^2 + \dfrac{x_m-\beta_3(1+U_f)}{\beta_1(1+U_f)}
\end{align}
Again, using $\beta_1>0$ from Remark \ref{rem:fx} and \eqref{eq:quad_low2},
\begin{align}
     &C_{1l}+C_{2l}x + C_{3l}x^2\geq C_{0l}^2, ~~ \forall x_m\in [\beta_3(1+U_f),\infty). \label{eq:quad_low3}
\end{align}
Using \eqref{eq:quad_low3}, one of the roots of \eqref{eq:quad_low1} is positive, that is,
\begin{align}
\begin{aligned}
     x_l&=C_{0l} + \sqrt{C_{1l}+C_{2l}x + C_{3l}x^2}, ~ ~x_m\in [\beta_3(1+U_f),\infty)  
\end{aligned}
\end{align}
where, using \eqref{eq:reldepth},
\begin{align}\label{eq:quad_low4}
\begin{aligned}
    x_m&\in [\beta_3(1+U_f),\infty) \\& \implies  x \in \left[\frac{-(\beta_2+1) + \sqrt{(\beta_2+1)^2-4 \beta_1\beta_3 U_f}}{2 \beta_1},\infty \right).
\end{aligned}
\end{align}
\end{proof}
\begin{remark}
    As governed by Eqs. \eqref{eq:up_depth}-\eqref{eq:coefflow}, the upper and lower bounds on $x$, that is, $x_u$ and $x_l$ explicitly depend on $x$.
\end{remark}

\subsection{Upper Bound on the Computed Closing Speed}\label{subsec}

The closing speed between the two vehicles can be computed using successive depth measurements. Accordingly, bounds on the computed depth affect the computed closing speed. In this section, an adaptive depth sampling logic is proposed to compute the bounds within which the closing speed between the ego vehicle and the neighboring vehicle lies while considering the uncertainty $[x_l,x_u]$ around the computed depth $x$.

 \subsubsection{Nominal closing speed}
\begin{figure}[!hbt]
     \centering
\includegraphics[width=\columnwidth]{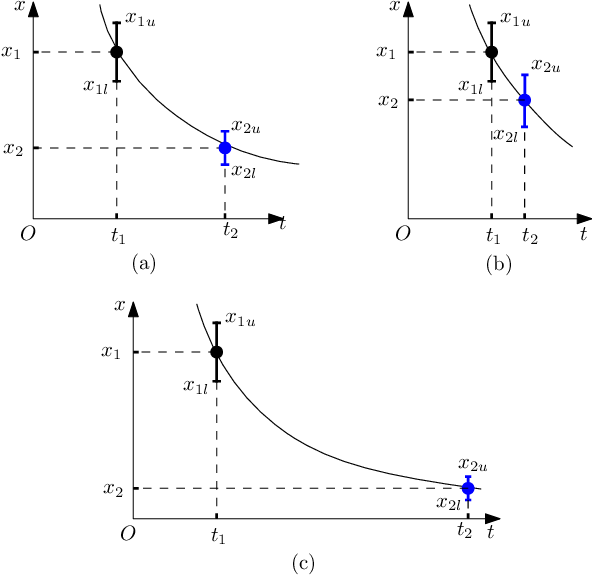}
     \caption{Computation of closing speed and its bounds when sampling distance is (a) nominal (b) very small (c) large.}
     \label{fig:velocity_com}
 \end{figure}

Let $x_1$ and $x_2$ ($x_1>x_2$) denote two successive computed depths of the neighboring vehicle at $t_1$ and $t_2$ ($t_2>t_1$), respectively, as shown in Fig. \ref{fig:velocity_com}(a). Let $\Delta x = (x_2-x_1)$ denote the sampling distance. The nominal closing speed is defined as the negative rate of change of the computed depth of the vehicle, that is,
\begin{equation}\label{eq:vnom}
    v_{\text{nom}} = - \dfrac{x_2-x_1}{t_2-t_1}  =- \dfrac{\Delta x}{\Delta t}.
\end{equation}

\subsubsection{Bounds on closing speed}
As shown in Fig. \ref{fig:velocity_com}(a), let the upper and lower bounds on the computed depth $x_k~(k = \{1,2 \})$ be $x_{ku}$ and $x_{kl}$, respectively, which leads the maximum relative displacement from $t_1$ to $t_2$ as $ (x_{2l}-x_{1u})$ and the minimum relative displacement as $(x_{2u}-x_{1l})$. Accordingly, the bounds on the nominal closing speed are calculated as 
\begin{equation}\label{eq:vu_vl}
    v_u =-\dfrac{x_{2l}-x_{1u}}{\Delta t},~v_l =-\dfrac{x_{2u}-x_{1l}}{\Delta t}
\end{equation}
where $v_u$ and $v_l$ are the upper and lower bounds, respectively. The upper bound on the closing speed is of specific relevance to the problem at hand and is further analysed.

\begin{remark}\label{rem:sampling}
Consider Fig. \ref{fig:velocity_com}(b) wherein the sampling distance is very small. Using \eqref{eq:vu_vl}, that leads to $v_u$ being unrealistically large as compared to $v_{\text{nom}}$. Conversely, a large $\Delta x$ depicted in Fig. \ref{fig:velocity_com}(c) results in a longer blind zone in which the closing speed is not updated.
\end{remark}

Remark \ref{rem:sampling} highlights the vital significance of the depth sampling rate in evaluating the closing speed. The following theorem directly relates the sampling distance to the upper bound on the computed closing speed.

\begin{theorem}\label{th:vel_bound}
Let the relative closing speed deviation be defined as $\gamma_u =  (v_u-v_{\text{nom}})/v_{\text{nom}}$. By imposing an upper threshold $\epsilon \in \mathbb{R}^{+}$ on $\gamma_u$, the sampling distance can be determined as
\begin{align}\label{eq:proposed_sd}
\begin{aligned} 
\Delta x  = x_2-x_1=\dfrac{C_{0l} + \sqrt{C_{1l}+C_{2l}x_2 + C_{3l}(x_2)^2} -x_{1u} }{1+\epsilon}       
\end{aligned}
\end{align}
\end{theorem}

\begin{proof}
Using \eqref{eq:vnom} and \eqref{eq:vu_vl}, $\gamma_u$ is given by 
\begin{align}\label{eq:up_vel_dev_init}
   \gamma_u = \dfrac{v_u - v_{\text{nom}} }{v_{\text{nom}}} = \dfrac{x_{1u}-x_{2l}- x_1 +x_2}{x_1-x_2}
\end{align}
Using \eqref{eq:low_depth}, substituting for the lower bound on $x_2$ in \eqref{eq:up_vel_dev_init},
\begin{align}\label{eq:gamma_u}
    \gamma_u = \dfrac{ x_2 -  C_{0l} - \sqrt{C_{1l}+C_{2l}x_2 + C_{3l}(x_2)^2}  +x_{1u} - x_{1}}{x_1 -x_2}
\end{align}
Imposing $\gamma_u = \epsilon$ in \eqref{eq:gamma_u},
\begin{align}\label{eq:x2_up}
 \dfrac{ x_2 - C_{0l} - \sqrt{C_{1l}+C_{2l}x_2 + C_{3l}(x_2)^2}  +x_{1u} -x_1  }{x_1 -x_2}=\epsilon.
\end{align}
Solving for $x_2$ in \eqref{eq:x2_up}, we obtain the sampling distance as $\Delta x = $ RHS of \eqref{eq:proposed_sd}.
\end{proof}

\section{Proposed Conflict Resolution Approach}\label{sec:conflict}

A second-order B\'ezier curve joining the points $P_i$ and $P_f$ in Fig. \ref{fig:prob_scenario} is considered as a path planning solution for performing the lane-exit maneuver. Further, based on the convex hull of the control points of the B\'ezier curve, an algorithm is presented for evaluating and avoiding conflicts with the neighboring vehicle.

\subsection{B\'ezier Curve Path Planning}
A second-order B\'ezier curve can be expressed as
\begin{equation}\label{eq:bezier_curve}
X_b(\tau) = 
    \begin{bmatrix}
           {x_b}(\tau) \\
           {y_b}(\tau) \\
         \end{bmatrix} = (1-\tau)^2P_i+ 2\tau(1-\tau)P_{int}+ \tau^2 P_f
\end{equation}
where $\tau\in [0,1]$ is the B\'ezier curve parameter, and $(P_i,P_f,P_{int}) \in \mathbb{R}^2$ are the control points of the B\'ezier curve. Following are some of the relevant properties of the B\'ezier curve:
\begin{enumerate}
    \item The curve lies within the convex hull formed using its control points.
    \item The curve passes through $P_i$ and $P_f$ but not through $P_{int}$.
    \item Tangents to the curve at $P_i$ and $P_f$ are along the line joining $P_i$ and $P_{int}$, and $P_f$ and $P_{int}$, respectively.
\end{enumerate}
In this work, $P_i = (x_i,y_i)^T$ and $P_f=(y_f,y_f)^T$ are known parameters. For the path tangents at the end points $P_i$ and $P_f$ to be along the respective lanes, $P_{int} = (x_{int},y_{int})^T$ can be uniquely determined as
\begin{align}
    x_{int}&= 
     \frac{y_f - y_i + x_i \tan(\theta_i) - x_f \tan(\theta_f)}{\tan(\theta_i) - \tan(\theta_f)}, \\
     y_{int}&=y_i + \left(x_{int} - x_i\right) \tan(\theta_i) .
\end{align}
where $\theta_i$ and $\theta_f$ are the orientation angles of Lanes 1 and 2, respectively. In the specific geometry shown in Fig. \ref{fig:algo1}, $\theta_i=0$ and $\theta_f=\pi/2$.

\subsection{Proposed Conflict Resolution Approach}

A decision logic is proposed to resolve any possible conflict between the two vehicles as they approach the T-intersection. The logic is presented as Algorithm 1 and details are discussed subsequently. The vehicles are assumed to be of identical dimensions with $l_s$ and $ w_s$ being their lengths and widths, respectively. While devising a conflict-free solution for the ego vehicle, the dimensions of the neighboring vehicle and the bounds on the computed depth $(x_l,x_u)$ are important factors. Accordingly, the obstacle region corresponding to the neighboring vehicle is defined as a rectangular region $\mathcal{N}_s$,
\begin{align}\label{eq:safe_zone}
\mathcal{N}_s = \bigg\{ (x, y) \ \bigg| \ &
p_E + \mathcal{R}_{\alpha}(\alpha_E)
\begin{pmatrix} 
x_l - \frac{l_s}{2} \\ y_m - \frac{w_s}{2} 
\end{pmatrix} \leq 
\begin{pmatrix}
x \\ y 
\end{pmatrix} \notag \\ 
& \leq 
p_E + \mathcal{R}_{\alpha}(\alpha_E) 
\begin{pmatrix} 
x_u + \frac{l_s}{2} \\ y_m + \frac{w_s}{2} 
\end{pmatrix} 
\bigg\}.
\end{align}
The time taken by the ego vehicle to traverse the exit path joining $P_i$ and $P_f$ in \eqref{eq:bezier_curve} can be computed numerically as
\begin{align}
    t_{\mathcal{C}} = \dfrac{\int\limits_{0}^{1}\mid \mid  -2(1-\tau)P_i+2(1-2\tau)P_{int} +2\tau P_f\mid \mid d\tau}{V_E}
\end{align}

\begin{figure}[!hbt]
    \centering
    \includegraphics[width=\columnwidth]{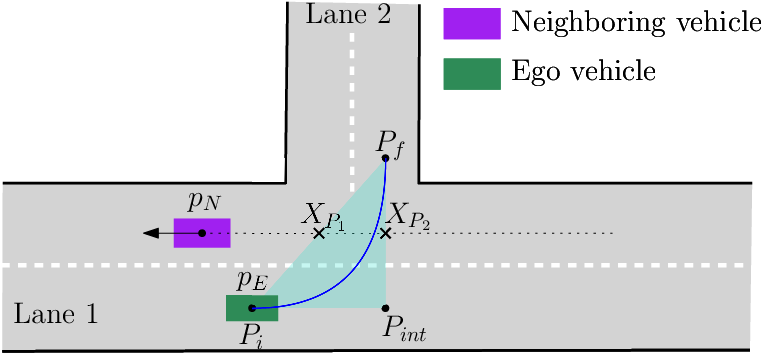}
    \caption{No conflict scenario: The neighboring vehicle has already crossed the intersection.}
    \label{fig:algo1}
\end{figure}
\begin{figure}[!hbt]
    \centering
    \includegraphics[width=\columnwidth]{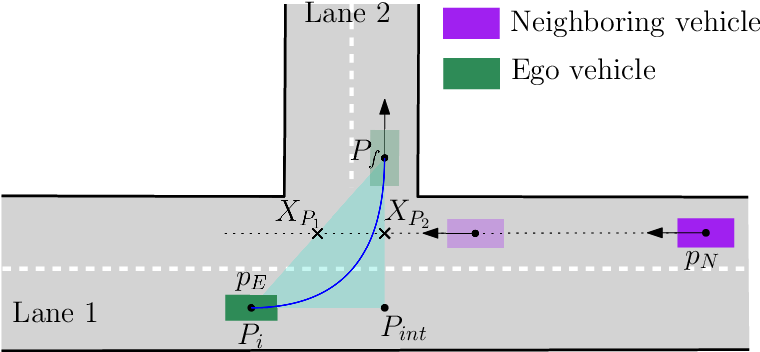}
    \caption{No conflict scenario: The neighboring vehicle is yet to reach the intersection when the ego vehicle reaches $P_f$. (Solid and faded boxes show vehicles' positions at $t_{P_i}$ and $t_{P_f}$, respectively.)}
    \label{fig:algo2}
\end{figure}

\vspace{0.5cm}
The subsequent discussions consider a local reference frame with its origin at $P_i$. Let $\mathcal{C}$ denote the convex hull formed using $P_i,~P_{int}$ and $P_f$. As shown in Fig. \ref{fig:algo1}, consider the points $X_{P_1}=(x_{p1},y_{p1})$ and $X_{P_2}=(x_{p2},y_{p2})$ along the centerline of the lane in which the neighboring vehicle is moving, which are obtained from the intersection of the segments $\overline{P_{i}P_f}$ and $\overline{P_{int}P_f}$, respectively, with the centerline of the lane where the neighboring vehicle is moving. Consider that at $t=t_{P_i}$, we have $p_E = P_i$. The scenario in Fig. \ref{fig:algo1} corresponds to $\mathcal{N}_s(t_{P_i}) \cap \mathcal{C} =\emptyset $ and $x_N(t_{P_i})<x_{p1}-d_s$ (line \ref{alg:line5} in Algorithm \ref{algo:conflict}), and this is considered as no conflict as the neighboring vehicle has crossed the T-intersection and is moving away from $X_{P_1}$. Accordingly, the decision variable $D_{v1}$ is assigned 1 and the ego vehicle starts following $X_b$ at $t =t_{P_i} $ (line \ref{alg:line8} in Algorithm \ref{algo:conflict}).

 \begin{algorithm}[!hbt]
\caption{Conflict Resolution for Lane-Exit Scenario}
\label{algo:conflict}
\textbf{Input: ${p}_N,p_E,P_i,P_{int},P_f,d_s$} \newline
\textbf{Output:} $D_{v1},D_{v2}$
\begin{algorithmic}[1]
\STATE Generate B\'ezier curve $X_b$ and convex hull $\mathcal{C}$ using the control points $(P_i,P_{int},P_f)$.
\STATE Compute $\mathcal{N}_s$ from the computed depth value using \eqref{eq:safe_zone} and Proposition \ref{proposition:bounds}.
\STATE Obtain the point $X_{P1}$ and $X_{P2}$.
\STATE Initialise two decision variables $D_{v1}$ and $D_{v2}$.
\STATE $D_{v1}\gets \text{if } (\text{at $t_{P_i},$ }\mathcal{N}_s\cap \mathcal{C}=\emptyset$ and $x_N<x_{p1}-d_s)$ then 1 else -1.\label{alg:line5}
\STATE $D_{v2}\gets$ if (at $t_{P_i}$ and $t_{P_f}$, $\mathcal{N}_s\cap \mathcal{C}=\emptyset$ and $x_N>x_{p2}+d_s$) then 1 else -1. \label{alg:line6}
\WHILE{$D_{v1}=-1$ and $D_{v2}=-1$}
        \STATE Wait at $P_i$. \label{alg:line7}
        \STATE Update $x_N,t_{P_i}$ and $t_{P_f}$.
        \STATE Compute $D_{v1}$ and $D_{v2}$.
\ENDWHILE
\STATE Execute $X_b$.\label{alg:line8}
\end{algorithmic}
\end{algorithm}

 \begin{figure*}[!hbt]
        \subfloat[]{%
            \includegraphics[width=.495\linewidth]{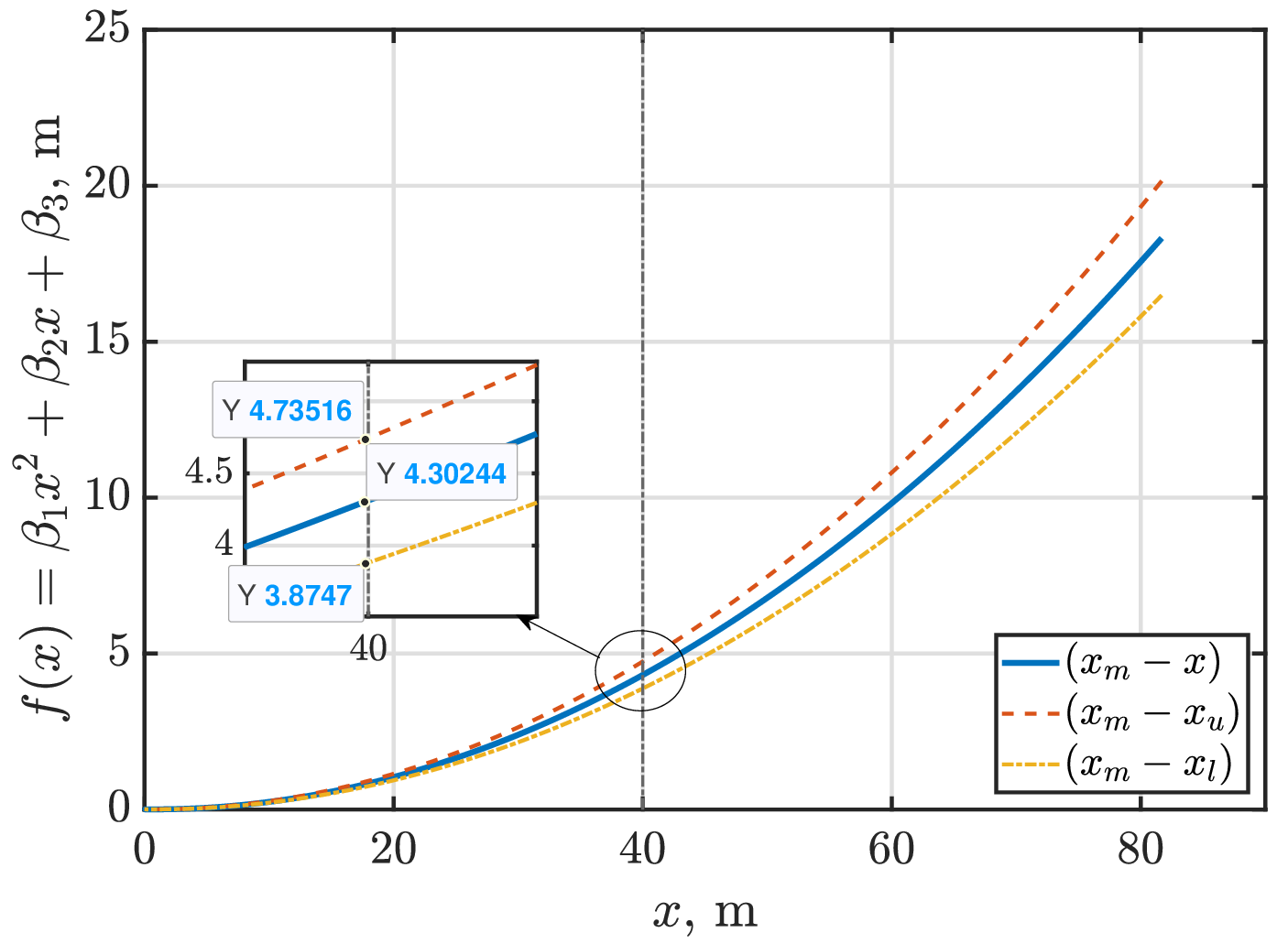}%
        }\hfill
        \subfloat[]{%
            \includegraphics[width=.495\linewidth]{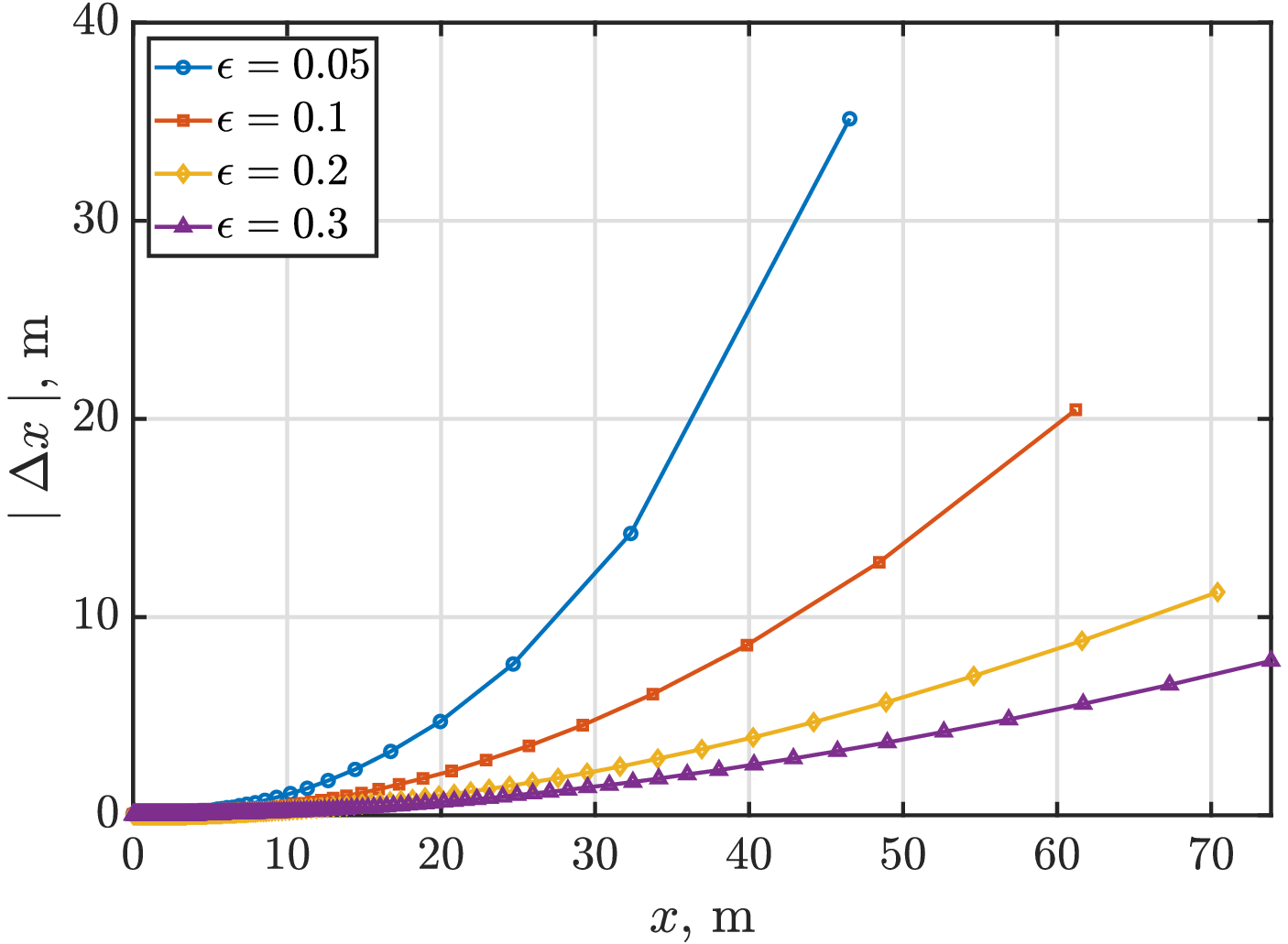}%
        }\hfill
        \subfloat[]{%
            \includegraphics[width=.495\linewidth]{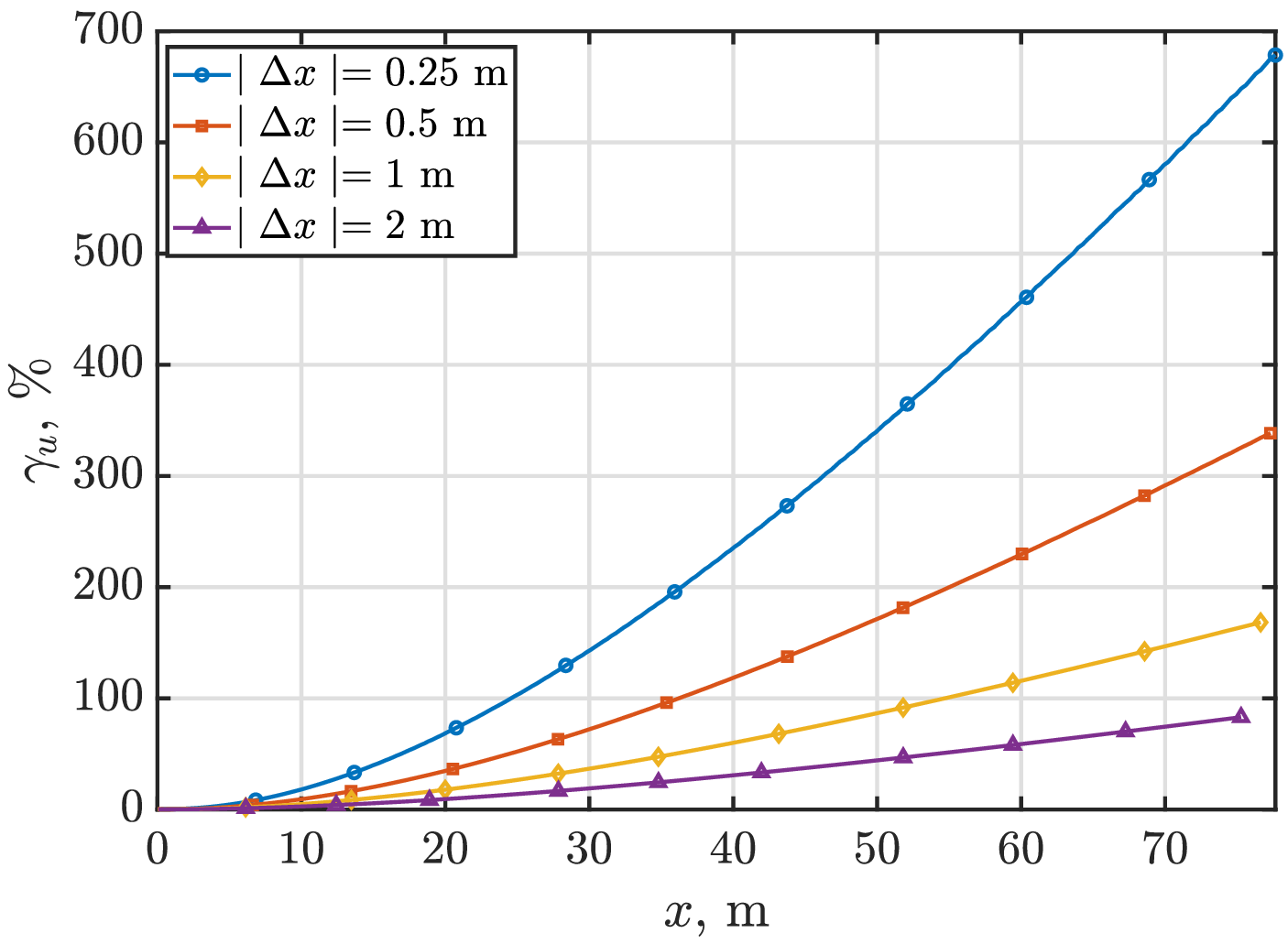}%
        }\hfill
        \subfloat[]{%
            \includegraphics[width=.495\linewidth]{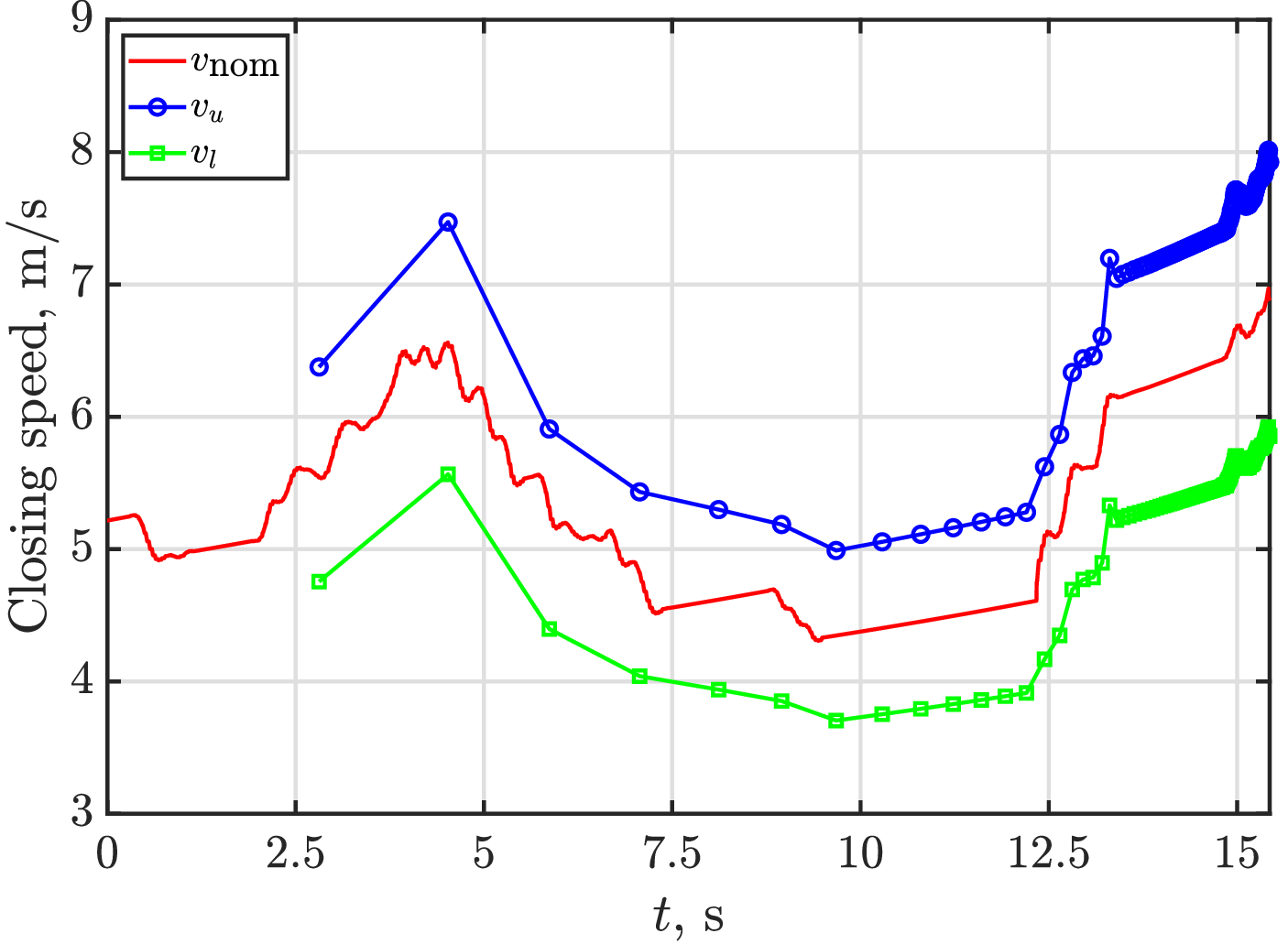}%
        }
        \caption{(a) Depth error variation with computed depth using $\beta_1 =0.002797,~\beta_2=-0.004249,~\beta_3=0.007311$. (b) Proposed sampling distance for abiding by a given relative deviation from the nominal closing speed. (c) Relative deviation from nominal closing speed considering fixed sampling distance $\Delta x$. (d) Nominal closing speed and its bounds considering $\epsilon=0.2$.}
        \label{fig:speed}
    \end{figure*}
    
The second scenario as shown in Fig. \ref{fig:algo2} considers the position of vehicles computed at $t_{P_f} =t_{P_i}+t_{\mathcal{C}}$ where initially at $t = t_{P_i}$, $\mathcal{N}_s\cap \mathcal{C}=\emptyset$ and $x_N>x_{p2}+d_s$. Considering the upper bound on the neighboring vehicle's closing speed, we compute $x_N(t_{P_f}) = x_N(t_{P_i}) -(v_u +V_E)t_{\mathcal{C}}$. Further, if at $t = t_{P_f}$, $\mathcal{N}_s\cap \mathcal{C}=\emptyset $ and $x_N(t_{P_f})>x_{p2}+d_s$, then $\mathcal{N}_s\cap \mathcal{C}=\emptyset,~\forall t \in [t_{P_i},t_{P_f}] $ (line \ref{alg:line6} in Algorithm \ref{algo:conflict}). Accordingly, the decision variable ${D}_{v2}$ is assigned 1 and the ego vehicle can safely execute the planned path $X_b$. All other possibilities apart from the two previously discussed scenarios are considered as conflict. In conflict scenarios, the ego vehicle waits at $P_i$ (line \ref{alg:line7} in Algorithm \ref{algo:conflict}) until the conditions, $\mathcal{N}_s \cap \mathcal{C} =\emptyset$ and $x_N(t_{P_i})<x_{p1}$ (Fig. \ref{fig:algo1}), are satisfied.

\section{Simulation Studies}\label{sec:simulation}
 In this section, we carry out studies validating the depth sampling logic developed in this work for closing speed computation, and the effectiveness of the proposed conflict resolution strategy in a dynamic traffic scenario at a T-intersection. The coefficients of the depth measurement error function in \eqref{eq:error_model} are based on the experiments conducted in \cite{cabrera2018versatile} as $\beta_1 =0.002797,~\beta_2=-0.004249,~\beta_3=0.007311$. The coefficient of determination for the curve fitting function $f(x)$ is considered as $R^2 = 0.9$. Vehicle dimensions are considered as $l_s = 3.8$ m, $w_s=1.8$ m. The minimum safety separation in \eqref{eq:ds} is considered as $d_s = 3.8$ m.

 
\begin{figure*}[!hbt]
  \centering
  \subfloat[ $t = 2.34$ s]{\includegraphics[width=\columnwidth,trim={1.45cm 0.07cm 2.1cm 1.50cm},clip]{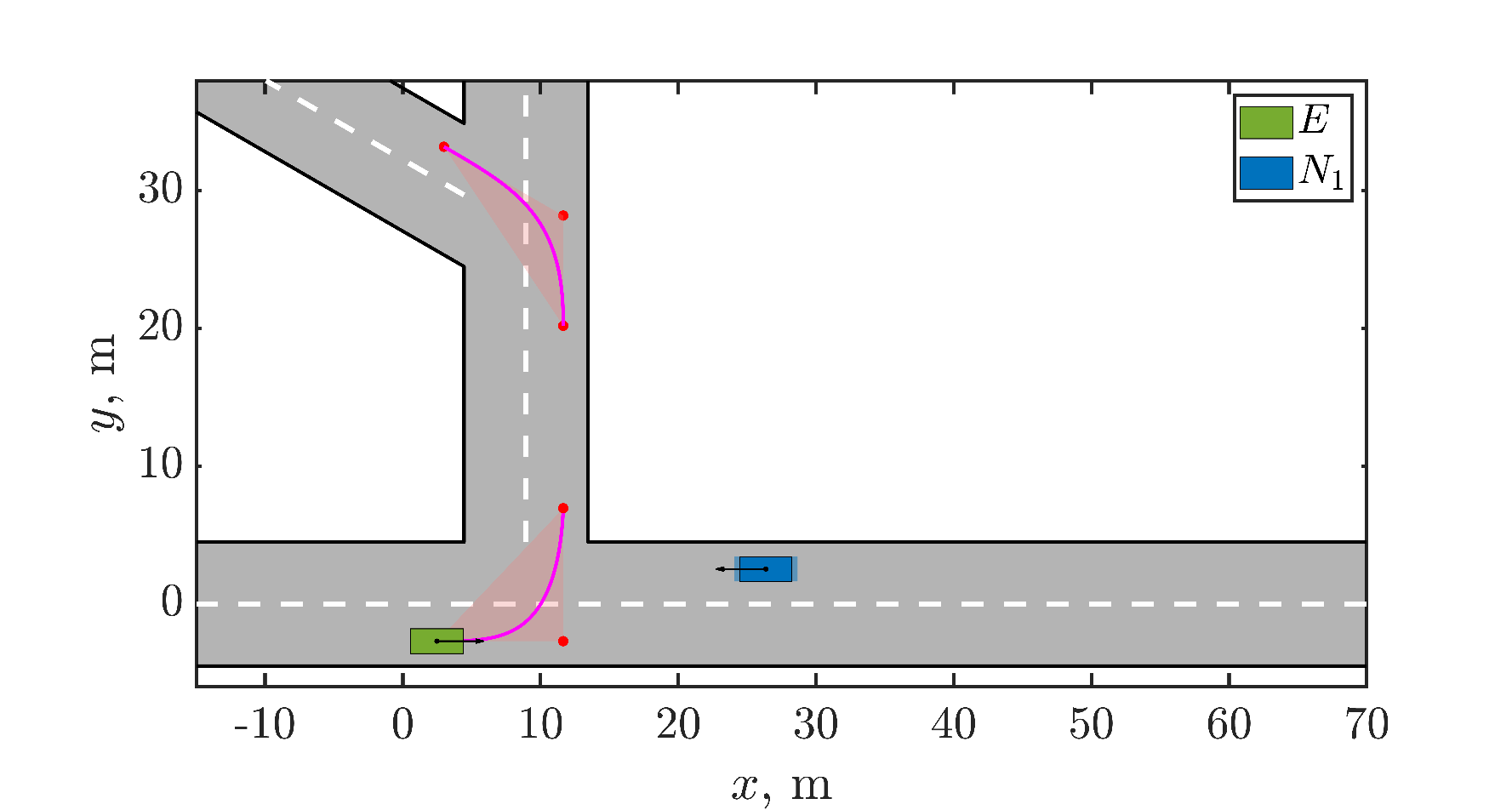}}\hfill
  \subfloat[$t = 7.39$ s ]{\includegraphics[width=\columnwidth,trim={1.45cm 0.07cm 2.1cm 1.50cm},clip]{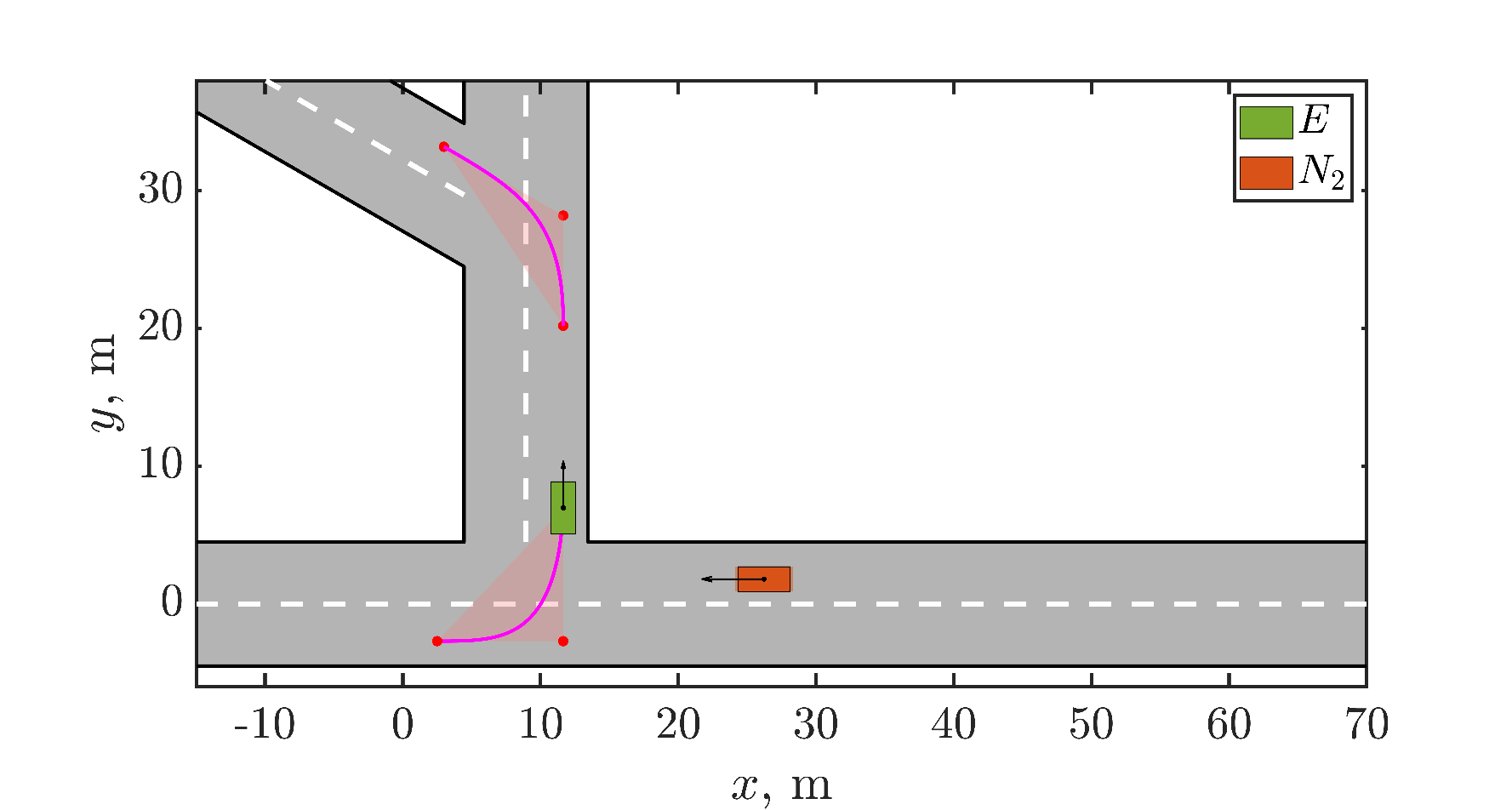}}\\
  \subfloat[$t = 9.20$ s ]{\includegraphics[width=\columnwidth,trim={1.45cm 0.07cm 2.1cm 1.50cm},clip]{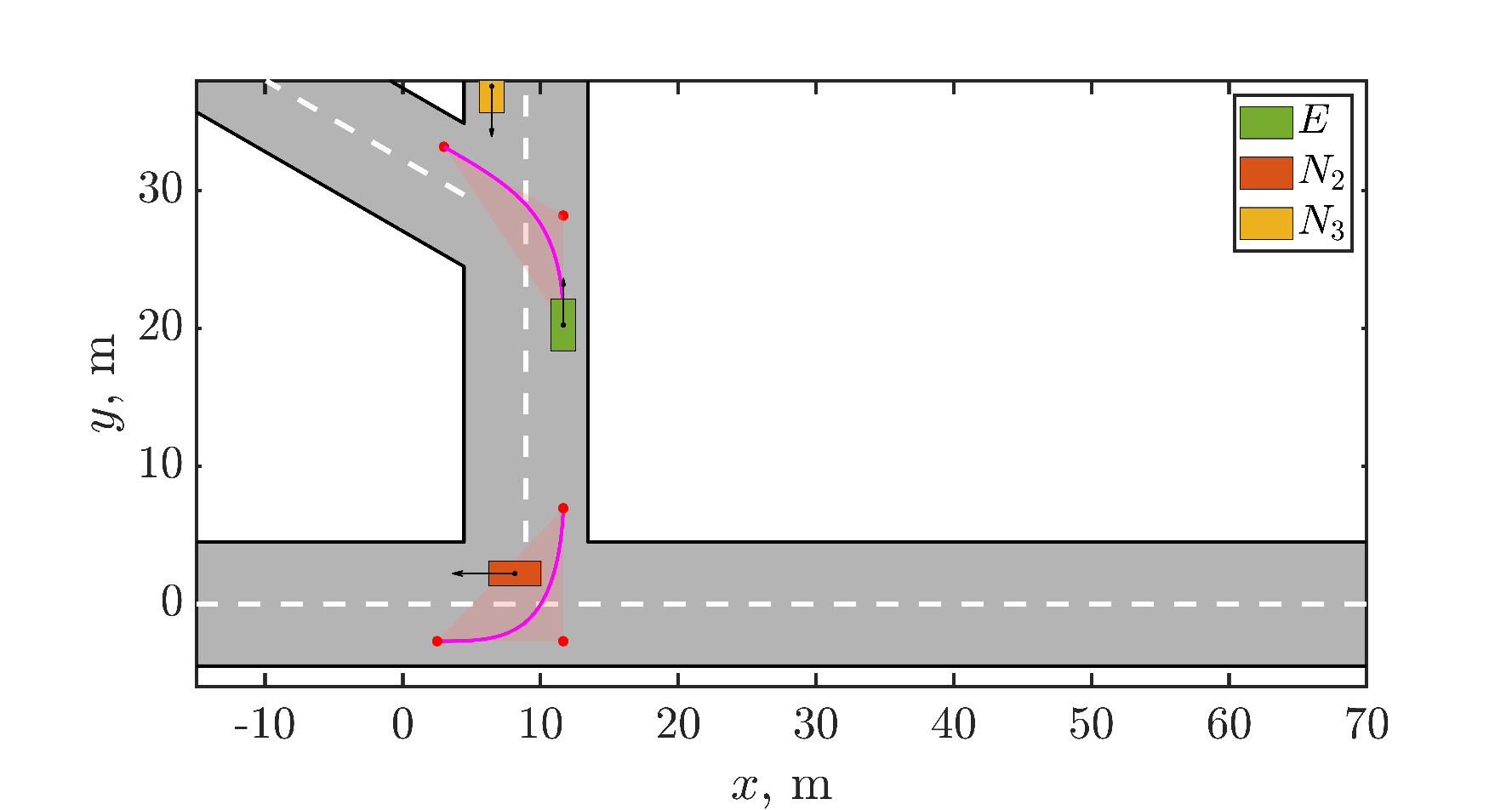}}\hfill
  \subfloat[$t = 12.05$ s ]{\includegraphics[width=\columnwidth,trim={1.45cm 0.07cm 2.1cm 1.50cm},clip]{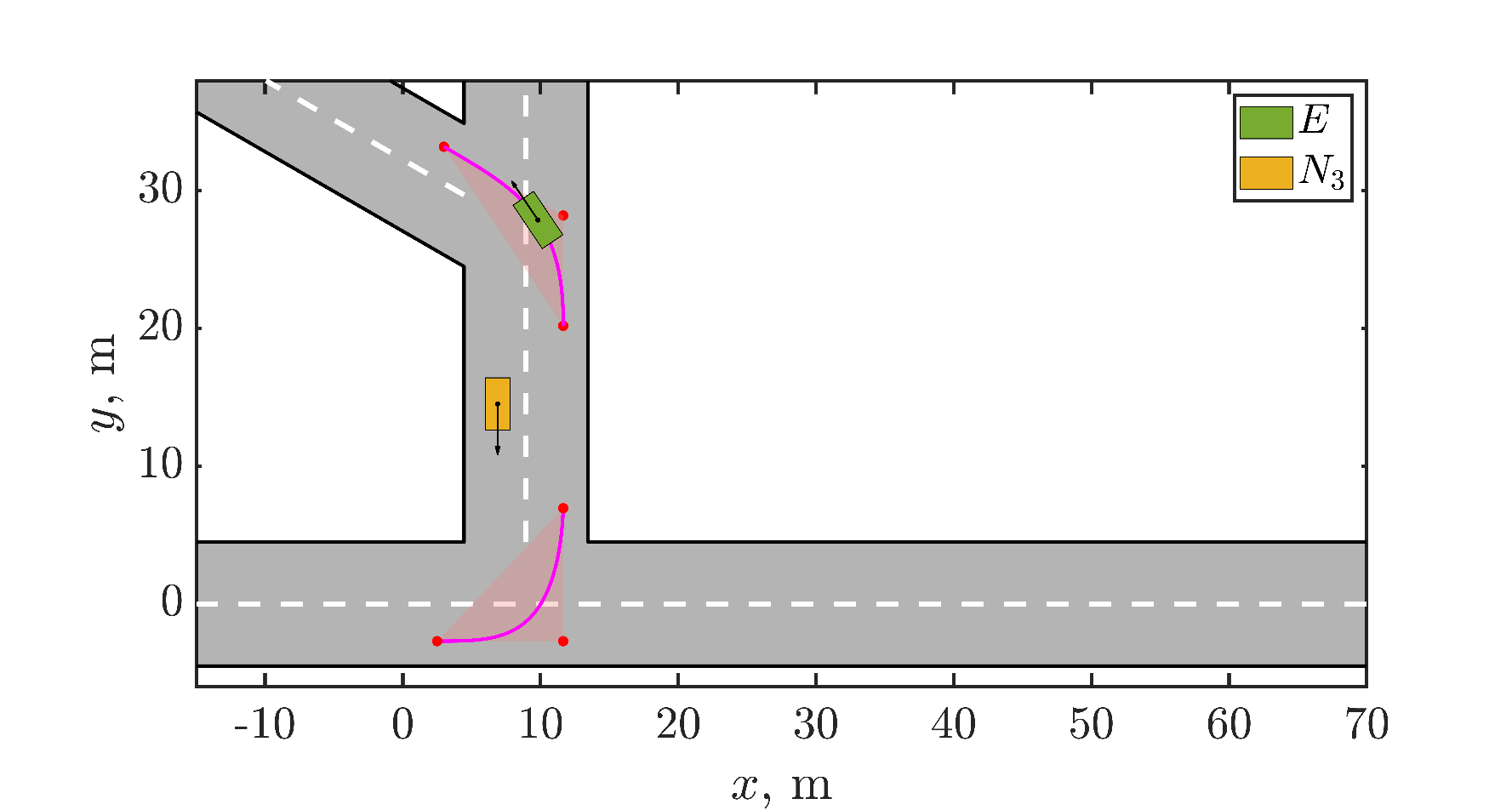}}
  \caption{Validation of conflict resolution algorithm in multi-vehicle dynamic traffic scenario (The animation video can be found \href{https://drive.google.com/file/d/1_3hUMrBhVBZ3j82cC5RjaLVUijfHQaZ5/view?usp=drive_link}{here}).}%
    \label{fig:animate}
\end{figure*}
Further, NGSIM dataset \cite{NGSIM} is used to realistically model neighboring vehicles' trajectories in its designated lane. The dataset consists of vehicle trajectories collected using 8 synchronized video cameras mounted at the top of a 36-story building. That study was done over a road segment of approximately 640 meters for a total of 45 minutes segmented into three intervals, each of 15 minutes duration, at different times of the day. The position information from the NGSIM dataset is used as the neighbouring vehicles' measured relative position $(x_m,y_m)$.

\begin{figure}
    \centering
    \includegraphics[width=\columnwidth]{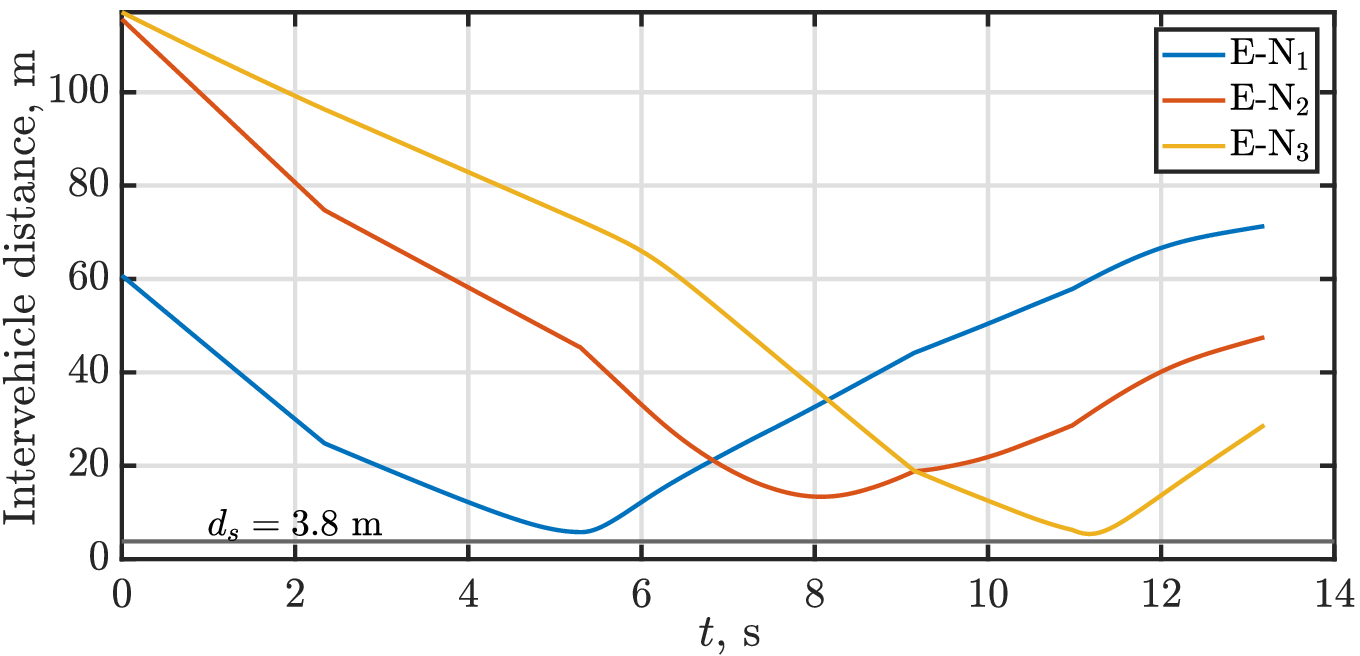}
    \caption{Distance between the neighboring vehicles and ego vehicle.}
    \label{fig:int_veh}
\end{figure}

\subsection{Closing Speed Computation Results}

As an example, a sample vehicle position data is selected from the NGSIM dataset wherein $x_m=100$ m is the initial measured depth of the neighboring vehicle. With this, the bounds on the computed depth and closing speed, and their effects on sampling distance are analysed subsequently. Using \eqref{eq:error_model} and the choice of ($\beta_1,\beta_2,\beta_3$), the variation in depth measurement error with the computed depth is shown in Fig. \ref{fig:speed}(a). As a sample point at $x =40$ m, an uncertainty of approximately $\pm43$ cm is observed in the curve fit. Shown in Fig. \ref{fig:speed}(b) is the absolute value of the sampling distance $\Delta x$ governed by \eqref{eq:proposed_sd} for maintaining a specified threshold $\epsilon$ on $\gamma_u$. Therein, for a smaller choice of $\epsilon$, the sampling distance is found to be sufficiently large. This is attributed to the fact that a lower deviation in computed closing speed necessitates the difference in successive computed depths to be significantly higher than the errors around them. Alternatively, sampling can be faster, that is, $\mid \Delta x \mid$ is small if $ \epsilon$ is relaxed to a higher value. Further, it can be noticed that the proposed sampling distance reduces as the depth reduces. This trend is attributed to reduction in position uncertainty with reducing depth as evident in Fig. \ref{fig:speed}(a). In contrast, as shown in Fig. \ref{fig:speed}(c), utilizing fixed sampling distance, the relative deviation of the upper bound on the closing speed with respect to the nominal closing speed increases drastically with increasing depth, rendering it impractical for decision-making. For the specific case of imposing $\epsilon$ = 0.2 using the proposed depth sampling logic, the nominal closing speed, and its upper and lower bounds are shown in Fig. \ref{fig:speed}(d). The markers, therein, represent the sampling instants.


 \subsection{Conflict Resolution Validation}

While Algorithm \ref{algo:conflict} considers one neighboring vehicle, a sequential use of the proposed algorithm is demonstrated as the ego vehicle traverses two T-intersections and encounters a total of 3 neighboring vehicles. The sampling distance is deduced by considering $\epsilon=0.2$ in \eqref{eq:proposed_sd}, which is then utilized for computing the closing speed and its upper bound using \eqref{eq:vnom} and \eqref{eq:vu_vl}, respectively.  The points $(P_i^1,P_{int}^1,P_f^1)$ for the first T-intersection are $(2.5,-2.7)$ m, $(11.65,-2.7)$ m and $(11.65,6.95)$ m, respectively, while those of the second T-intersection, $(P_i^2,P_{int}^2,P_f^2)$ are $(11.65,20.2)$ m, $(11.65,28.2)$ m and $(2.98,33.2)$ m, respectively. While the ego vehicle is moving, its forward velocity $V_E= 7$ m/s. Using \eqref{eq:bezier_curve}, the heading angle input of the ego vehicle is governed by
\begin{align}
    \alpha_E(\tau) = \dfrac{\text{d}y_b/\text{d}\tau}{\text{d}x_b/\text{d}\tau}=\dfrac{-2(1-\tau)y_i+2(1-2\tau)y_{int} +2\tau y_f}{-2(1-\tau)x_i+2(1-2\tau)x_{int} +2\tau x_f}
\end{align}
Fig. \ref{fig:animate} shows the positions of the vehicles at different time instants. The ego vehicle, denoted as $E$ in Fig. \ref{fig:animate}, reaches $P_i^1$ at $t = 2.34$ s. At this instant, a conflict is detected with the vehicle $N_1$ and the ego vehicle waits at $P_i^1$ until $t = 5.27$ s when the conditions in line \ref{alg:line5} in Algorithm \ref{algo:conflict} are satisfied. At $t=5.27$ s, using line \ref{alg:line6} in Algorithm \ref{algo:conflict}, the ego vehicle is determined to have no conflict with the vehicle $N_2$ approaching intersection-1. Accordingly, the ego vehicle executes the lane exit maneuver between $t =5.2$ s to $7.39$ s, and subsequently reaches $P_i^2$ at $t = 9.20$ s. The ego vehicle further waits at $P_i^2$ from $t=9.20$ s to $10.98$ s as a conflict with the vehicle $N_3$ is detected. At $t=10.98$ s, $N_3$ moves across the second T-intersection, and at that instant, the ego vehicle initiates the lane-exit maneuver. The distance between neighboring vehicles and the ego vehicle in Fig. \ref{fig:int_veh} shows that the distances are throughout higher than $d_s$.

\section{Conclusion}\label{sec:conclusion}

This letter presents a decision-cum-planning strategy considering uncertain depth information obtained using a stereo camera for an autonomous vehicle performing a lane-exit maneuver at a T-intersection. Considering a quadratic error model for the depth measurements and an associated coefficient of determination, the nominal depth and its bounds are computed. By imposing an upper limit on the deviation in the relative closing speed, a relationship between the sampling distance, and the nominal depth and its bounds is established. A quadratic B\'ezier curve is considered as a path planning solution for the ego vehicle performing the lane-exit maneuver. Based on the upper bound on the closing speed and the convex hull of the control points of the B\'ezier curve, a decision logic is presented for the ego vehicle to safely execute the lane-exit maneuver at T-intersections. Validation studies are carried out in a dynamic traffic scenario with multiple neighboring vehicles whose trajectories are modelled using the realistic NGSIM dataset.

Future works could be towards studying closing speed evaluation frameworks for different sensors like lidar and radar, and extending the approach in different traffic scenarios. 

\section*{Acknowledgement}

The effort was supported by WIPRO-IISc Research and Innovation Network (WIRIN) via grant no. 99325W.

\bibliographystyle{IEEEtran} 
\bibliography{sample}

\end{document}